\documentclass[letterpaper, 10 pt, conference]{ieeeconf}  

\IEEEoverridecommandlockouts                              

\usepackage{graphics} 
\usepackage{epsfig} 
\usepackage{amsmath} 
\usepackage{amssymb}  

\usepackage{caption}
\usepackage{subcaption}

\usepackage{xspace}
\newcommand*{\eg}{e.g.\@\xspace}
\newcommand*{\ie}{i.e.\@\xspace}

\title{\LARGE \bf
Optimal Whole Body Trajectory Planning for Mobile Manipulators in Planetary Exploration and Construction
}

\author{Federica Storiale$^{1}$, Enrico Ferrentino$^{1}$, Federico Salvioli$^{2}$, Konstantinos Kapellos$^{3}$ and Pasquale Chiacchio$^{1}$
\thanks{$^{1}$Authors are with the Department of Information and Electrical Engineering and Applied Mathematics (DIEM), University of Salerno, Fisciano, SA, 84084, Italy,
        {\tt\small \{fstoriale,eferrentino, pchiacchio\}@unisa.it}}%
\thanks{$^{2}$Federico Salvioli is with Robotic Exploration, ALTEC S.p.A., 10146 Torino, Italy,
        {\tt\small federico.salvioli@altecspace.it}}%
\thanks{$^{3}$Konstantinos Kapellos is with Trasys International, 3001 Heverlée, Belgium, 
        {\tt\small konstantinos.kapellos@trasysinternational.com}}%
}

\renewcommand{\baselinestretch}{0.978}

\begin{document}

\maketitle
\thispagestyle{empty}
\pagestyle{empty}

\begin{abstract}

Space robotics poses unique challenges arising from the limitation of energy and computational resources, and the complexity of the environment and employed platforms. At the control center, offline motion planning is fundamental in the computation of optimized trajectories accounting for the system's constraints. Smooth movements, collision and forbidden areas avoidance, target visibility and energy consumption are all important factors to consider to be able to generate feasible and optimal plans. When mobile manipulators (terrestrial, aerial) are employed, the base and the arm movements are often separately planned, ultimately resulting in sub-optimal solutions. We propose an Optimal Whole Body Planner (OptiWB) based on Discrete Dynamic Programming (DDP) and optimal interpolation. Kinematic redundancy is exploited for collision and forbidden areas avoidance, and to improve target illumination and visibility from onboard cameras. The planner, implemented in ROS (Robot Operating System), interfaces 3DROCS, a mission planner used in several programs of the European Space Agency (ESA) to support planetary exploration surface missions and part of the ExoMars Rover's planning software. The proposed approach is exercised on a simplified version of the Analog-1 Interact rover by ESA, a 7-DOFs robotic arm mounted on a four wheels non-holonomic platform.

\end{abstract}

\section{Introduction} \label{sect:introduction}

In the context of planetary colonization, space robots act as human precursors for exploration, terrain preparation and infrastructures construction \cite{Gao_2017}, performing activities otherwise unfeasible for men, allowing a reduction of risks and costs. Common scenarios include robotic arms mounted on rovers, that manipulate, transport, and assemble components for building infrastructures on the Moon or Mars, or on spacecrafts, for samples collection \cite{Lauretta_2017}. These applications require robots to move safely and efficiently in a challenging environment. Because of limited resources, special attention must be paid to energy and computational requirements. In this context, offline trajectory planning, \ie on ground, where the unit cost of computational power is lower, plays a fundamental role both in the mission design phase, for feasibility and budget assessments, and in the operations phase, to deliver efficient activity plans. Offline planning allows engineers to tailor trajectories for each task according to optimality criteria, taking into account the robot's and the environment's constraints. For example, it ensures that the robot moves smoothly, accurately, and efficiently to the desired location, while avoiding collisions and non-traversable areas. This is essential to prevent damage to the robot and to ensure the integrity of collected samples and manipulated objects.

In space robotics, the planning problem is often decomposed: the trajectory for the mobile base is planned through sampling-based approaches \cite{Chitta_2012, Sustarevas_2021, Sustarevas_2022, Nagatani_2002} or dynamic programming-inspired algorithms \cite{Nguyen_2023}, while the arm motion is computed through inverse kinematics, spline interpolation, genetic algorithms, reachability analysis \cite{Sandakalum_2022}. The reason behind the choice of a decoupled approach is the increased complexity of whole-body planning, also in terms of computational resources \cite{Nguyen_2023}, as it requires managing multiple degrees of freedom and different constraints for the arm and the base at the same time. However, this separation can only generate sub-optimal plans \cite{Ghodsian_2023} and, in the worst case, compromise feasibility \cite{Sandakalum_2022}. On the contrary, whole-body motion planning \cite{Oriolo_2005, Fernandez_2022, Salvioli_2022, Mi_2019} allows exploiting the system's kinematic redundancy to accomplish secondary objectives, optimizing performance indices of interest. \textit{Discrete Dynamic Programming} (DDP) is a global optimization technique that is frequently employed when an efficient solution is needed to a non-convex trajectory planning problem \cite{Guigue_2007, Gao_2017_DP, Kaserer_2019}: it has been proven effective for mobile manipulators \cite{Salvioli_2022} and constrained systems in general \cite{Ferrentino_2019}. Nevertheless, its discrete nature yields non-smooth trajectories, whose conversion into actual commands for real systems remains questionable \cite{Salvioli_2022}.

In this paper, building upon the algorithm proposed in \cite{Salvioli_2022}, we design an \textit{Optimal Whole Body Planner} (OptiWB) preserving solution optimality through DDP and guaranteeing trajectory smoothness through optimal interpolation. We demonstrate the effectiveness of our approach in a real space exploration use case, where kinematic redundancy is exploited to improve camera visibility over a target that must be reached by the robotic arm. The whole-body approach of OptiWB allows generating efficient plans, while simplifying the planning process at the control center: the ground operator can generate complex synchronized movements of the mobile platform and the arm by solely providing the desired trajectory for the arm's end-effector. This procedure is faster and less error-prone than defining cumbersome maneuvers for the rover and the arm independently.

While OptiWB is implemented in ROS, which would make its future integration with Space ROS \cite{space-ros} effortless, we demonstrate the versatility of our planner by integrating it with the 3D Rover Operations Control Software (3DROCS) \cite{Kapellos_2014, 3DROCS_ESA}, a commercial mission planning system by Trasys International, already employed in missions of space exploration and colonization. The robotic platform is a simplified version of the Analog-1 Interact rover \cite{analog_1_interact_rover}, composed of a differential drive non-holonomic mobile base and a LBR iiwa 14 R820 7-DOF arm \cite{kuka}.

In Section \ref{sect:OptiWB}, OptiWB is presented, with a description of both the DDP-based planner and the optimal interpolation. In Section \ref{sect:constraints} and Section \ref{sect:objective_function}, the constraints and the objective function designed for the considered exploration use case are described, respectively. The integration with 3DROCS is discussed in Section \ref{sect:3DROCS}. The results of applying OptiWB to our use case are presented in Section \ref{sect:experimental_results}, while Section \ref{sect:conclusions} draws the conclusions of our work.

\section{Optimal Whole Body Planner (OptiWB)}\label{sect:OptiWB}

\begin{figure}[]
\centering
\includegraphics[width=0.48\textwidth]{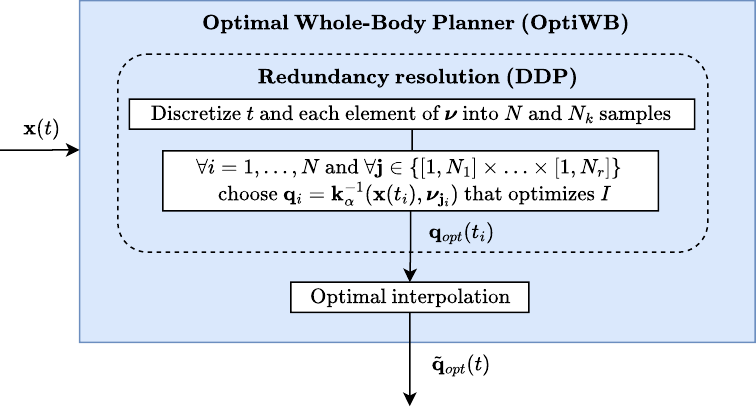}
\caption{\label{fig:OptiWB}Optimal Whole-Body Planner (OptiWB) architecture.}
\end{figure}

The high-level architecture of OptiWB is given in Fig.\ \ref{fig:OptiWB}. Each component is described next. 

\subsection{Discrete Dynamic Programming} \label{sect:ddp_planner}

Let ${\bf x}(t): \mathbb{R} \rightarrow \mathcal{X} \subseteq SE(3)$ be the desired task space trajectory with respect to time $t$, designed at the ground control center, either as a result of a higher-level task planner, or manually, by human operators, possibly by the means of graphical tools. Every ${\bf x}(t)$, for a fixed $t$, is termed a \emph{pose} and is characterized by $m$ degrees of freedom, which is the dimension of $\mathcal{X}$. The joint space commands (or joint space trajectory) ${\bf q}(t): \mathbb{R} \rightarrow \mathbb{R}^n$ for the robot are related to the task space trajectory, at the positional (zero-order) level, through \textit{inverse kinematics}, i.e.\
\begin{equation} \label{eq:inverse_kinematics}
        {\bf q}(t) = {\bf k}^{-1}\big({\bf x}(t)\big),
\end{equation}   
where ${\bf k}: \mathbb{R}^n \rightarrow SE(3)$ is the \textit{forward kinematics} function. When the number of joints $n$ is greater than the number of variables needed to describe the task $m$, the degree of redundancy is $r = n - m > 0$. 
For redundant robots, \eqref{eq:inverse_kinematics} admits an infinite number of solutions, that are joint configurations corresponding to the same task. The process of solving \eqref{eq:inverse_kinematics} for each $t$, when $r>0$, is termed \textit{redundancy resolution} and can be performed by optimizing some performance criterion.

The approach proposed in \cite{Ferrentino_2019} is based on redundancy parametrization \cite{Ferrentino_2019_2}, \ie the choice of a set of joint positions (or functions of them) $\boldsymbol{\nu} \in \mathbb{R}^{r}$ representing redundancy, such that \eqref{eq:inverse_kinematics}  becomes:
\begin{equation} \label{eq:augmented_ik}
	{\bf q}(t) = {\bf k}_{\alpha}^{-1}\big({\bf x}(t), \boldsymbol{\nu}(t)\big),
\end{equation}
where ${\bf k}_{\alpha}: \mathbb{R}^n \rightarrow \mathbb{R}^n$ is the forward kinematics ${\bf k}$, augmented with the redundancy parameters' equations. The mapping in \eqref{eq:augmented_ik} is square, hence, for fixed $\boldsymbol{\nu}$ and ${\bf x}$, the vector of joint positions ${\bf q}$ belongs to a finite set.

Considering systems composed of a mobile base and an arm, ${\bf q} = \left[ {\bf q}_b^T,  {\bf q}_a^T \right]^T$, where ${\bf q}_b \in \mathbb{R}^{n_b}$ is the base configuration, ${\bf q}_a \in \mathbb{R}^{n_a}$ is the arm configuration, and $n_a+n_b = n$. For terrestrial robots, for the sake of planning, we take $n_b = 3$ \cite{Salvioli_2022}, therefore we model the base's state as
\begin{equation} \label{eq:base_coordinates}
    {\bf q}_b = [x \quad y \quad h]^T,
\end{equation}
where $x$ and $y$ are the linear displacements of the base's center of mass (COM) with respect to some fixed frame $\mathcal{F}_f$ (e.g.\ the starting configuration), that are indeed equivalent to prismatic joints; $h$ is the base's angular displacement with respect to $\mathcal{F}_f$, i.e.\ the \textit{heading}, that is indeed equivalent to a revolute joint. The other degrees of mobility (altitude, pitch and roll) are constrained by the geometry of the terrain surface, that here, for the sake of simplicity, we assume to be flat.

In order to address the problem with DDP, $t$ is discretized into $N$ samples and each element of $\boldsymbol{\nu}$ into $N_k$ samples, with $k=0,\ldots,r$. Thus, the problem becomes an optimal path search one, on a grid in $t_i$ and $\boldsymbol{\nu}_{\bf j}$, where ${\bf j} \in \lbrace \left[ 1, N_1\right] \times \ldots \times \left[ 1, N_{r} \right] \rbrace \subset \mathbb{N}^{r}$ is the vector of redundancy parameter indices. In the resulting grid, the goal is to find ${\bf q}(t_i) = {\bf q}(i)$, that optimizes the global cost
\begin{equation} \label{eq:dp_cost_function}
    I(N) = l({\bf q}(0)) + \sum_{i=1}^{N} \phi\big({\bf q}(i), \dot{\bf q}(i)\big).
\end{equation}
In \eqref{eq:dp_cost_function}, $l({\bf q}(0))$ is the cost associated to the initial configuration and $\phi$ is the objective function to optimize along the trajectory, which, for simplicity, is assumed to be only a function of joint positions ${\bf q}$ and velocities $\dot{\bf q}$. Therefore, the optimization problem, in a recursive form, is
\begin{equation}
\begin{split} \label{eq:dp_optimization_problem}
I_{opt}(0) & = l({\bf q}(0)), \\
I_{opt}(i) & = \min_{{\bf q}(i-1) \in \mathcal{B}(i-1)} \Bigr[\phi\big({\bf q}(i), \dot{\bf q}(i)) + I_{opt}(i-1) \Bigl],
\end{split}
\end{equation}
where $\mathcal{B}(i)$, for each $i$, is the feasible domain of ${\bf q}(i)$, and depends on the imposed constraints (see Section \ref{sect:constraints}); $I_{opt}$ is the optimal return function, hence $I_{opt}(N)$ is the final optimal cost. It is worth noting that, in \eqref{eq:dp_optimization_problem}, at each iteration, both $\dot{\bf q}(i)$, because of discrete approximations of derivatives, and $I_{opt}(i-1)$ depend on the choice made for ${\bf q}(i-1)$.

\subsection{Optimal interpolation} \label{sect:optimal_interpolation}

The joint space trajectory ${\bf q}_{opt}(t_i)$ provided by DDP is discretization-optimal, \ie it is the global optimum in the discrete domain defined by the indices $i$ and ${\bf j}$. To control the planning time, $N$ and $N_{k}$, i.e.\ the span of $i$ and ${\bf j}$, respectively, cannot be arbitrarily large. In turn, this yields non-smooth trajectories, compromising feasibility on real systems. To cope with this, we design a smoothing algorithm based on optimal interpolation. To impose the task constraint, such interpolation is performed in the domain of the spline control points associated with the redundancy parameters $\boldsymbol{\nu}(i)$. Once this is done, the other joint variables are retrieved through \eqref{eq:augmented_ik}. Thus, said $t_f$ the task duration, the optimization problem is formalized as follows:
\begin{align} \label{eq:optimal_smoothing_problem}
	\min_{\boldsymbol{\nu}(0), \ldots, \boldsymbol{\nu}({N})} & \int_{0}^{t_f} \phi(\tilde{\bf q}, \dot{\tilde{\bf q}}),\\
	s.t. \quad &{\bf q}(i) = {\bf k}_{\alpha}^{-1}({\bf x}(i), \boldsymbol{\nu}(i)) \quad \forall i=1,\ldots,N \label{eq:ik} \\ 
           &\tilde{\bf q}(t) = interp({\bf q}) \label{eq:interp_q_spline} \\ 
           & \tilde{\bf q}(t) \in \mathcal{B}(t) \quad \forall t\label{eq:derivative_constr}
\end{align}
where the objective function $\phi$ is evaluated on the interpolated trajectory $\tilde{{\bf q}}$;
the decision variables $\boldsymbol{\nu}(i)$, with $i = 1,...,N$, are initialized with the result of DDP;
inverse kinematics \eqref{eq:ik} exploits the decision variables $\boldsymbol{\nu}(i)$ to find the corresponding joint configurations at each $i$; the interpolation function $interp$ in \eqref{eq:interp_q_spline} embeds both linear interpolation for $x$, $y$ and $h$ and 3rd-order B-spline interpolation for all the other joints; \eqref{eq:derivative_constr} verifies that joint configurations respect first- and second-order derivatives embedded in the sets $\mathcal{B}(t)$, discussed in Section \ref{sect:constraints}.
With regards to the $interp$ function in \eqref{eq:interp_q_spline}, B-spline interpolation is preferred over simple polynomials because it allows for an easier manipulation of the curves \cite{Babaei_2018, Li_2022}. However, if applied to the base's degrees of freedom, it might lead to overshoots that would be difficult to control on a non-holonomic base. For this reason, in this work, linear interpolation is preferred for the base variables, resulting in turn-in-place maneuvers at non-differentiable points between linear segments. Nevertheless, more sophisticated interpolation functions may be adopted in \eqref{eq:interp_q_spline} without changing the general scheme.

We note that the objective function $\phi$ in \eqref{eq:optimal_smoothing_problem} is the same as DDP, as defined in \eqref{eq:dp_cost_function}, and, likewise, it is subject to integral optimization. The optimization problem \eqref{eq:optimal_smoothing_problem}-\eqref{eq:derivative_constr} is non-convex, therefore most well-known numerical optimization techniques can only guarantee local optimality. Nevertheless, in our strategy, DDP ``informs'' this local search by providing an optimal seed resulting from a global search, aiding both feasibility and optimality.

\section{Constraints formulation} \label{sect:constraints}

In our planning use case, we consider the following constraints to be satisfied along the entire trajectory: end-effector path, i.e.\ forward kinematics constraint; joint position, velocity and acceleration limits; avoidance of obstacles and forbidden areas; pure rolling constraint, assuming a non-holonomic base.
Then, at the final waypoint, we also consider the additional constraint that the target must not lie in the robot's shade, so as to guarantee the target illumination and improve its visibility from onboard cameras.

\subsection{Joint limits and pure rolling constraint}

Joint limits for the arm are formalized as
\begin{align}
\underline{{\bf q}_{a}} \le & {\bf q}_a \le \overline{{\bf q}_{a}}, \label{eq:joint_pos_limits} \\
\underline{\dot{{\bf q}}_{a}} \le & \dot{{\bf q}}_a \le \overline{\dot{{\bf q}}_{a}}, \label{eq:joint_vel_limits} \\
\underline{\ddot{{\bf q}}_{a}} \le & \ddot{{\bf q}}_a \le \overline{\ddot{{\bf q}}_{a}}, \label{eq:joint_acc_limits}
\end{align}
while, for the mobile base:
\begin{equation} \label{eq:base_joint_limits}
    \begin{split}
        \sqrt{\dot{x}^2 + \dot{y}^2} \leq v_{max},\\
        |\dot{h}| \leq \omega_{max},
    \end{split}
\end{equation}
where $v_{max}$ and $\omega_{max}$ are the maximum tangential and angular velocities of the base, respectively.

If a non-holonomic base is considered, the pure rolling constraint is
\begin{equation} \label{eq:pure_rolling_constr}
    \tan(h) = \frac{dy}{dx}.
\end{equation}

\subsection{Collision and forbidden areas avoidance}

Let ${\bf q}_p(t) = \left[ q_1(t), \ldots, q_p(t) \right]$ be the vector of joint positions up to the $p$-th joint at time $t$ and $V_p\big({\bf q}_p(t)\big)$ be the convex hull of the $p$-th link at same time. Let $V\big({\bf q}(t)\big)$ (without a subscript) be the non-connected volume occupied by the robot at time $t$, in the joint configuration ${\bf q}(t)$, formalized as the union of the single links' connected volumes, i.e.\
\begin{equation}
V\big({\bf q}(t)\big) = \bigcup_{p=1}^n V_p\big({\bf q}_p(t)\big).
\end{equation}
Let $V_e$ be the possibly non-connected volume of static environment obstacles. Collision and self-collision constraints are respectively formalized as
\begin{align}
    & V\big({\bf q}(t)\big) \cap V_e = \emptyset \quad \forall t, \label{eq:env_collision_constraints}\\
    & V\big({\bf q}_p(t)\big) \cap V\big({\bf q}_k(t)\big) = \emptyset \quad \forall t,p,k  = 1, \ldots, n \text{ with } p \neq k.\label{eq:self_collision_constraints}
\end{align}

Let $\mathcal{C}_{FA}$ be the set of $(x,y)$ coordinates corresponding to forbidden areas in a 2-D occupancy map, usually acquired by the means of visual systems. Then, the non-traversable areas avoidance constraint is verified as:
\begin{equation} \label{eq:forbidden_areas_constraint}
    (x(t),y(t)) \notin \mathcal{C}_{FA} \quad \forall t
\end{equation}
where $x(t)$ and $y(t)$, following their definition in \eqref{eq:base_coordinates}, are the coordinates of the mobile base at the time instant $t$.

\subsection{Overshadowing avoidance at the final waypoint}

\begin{figure}[]
\centering
\includegraphics[width=0.40\textwidth]{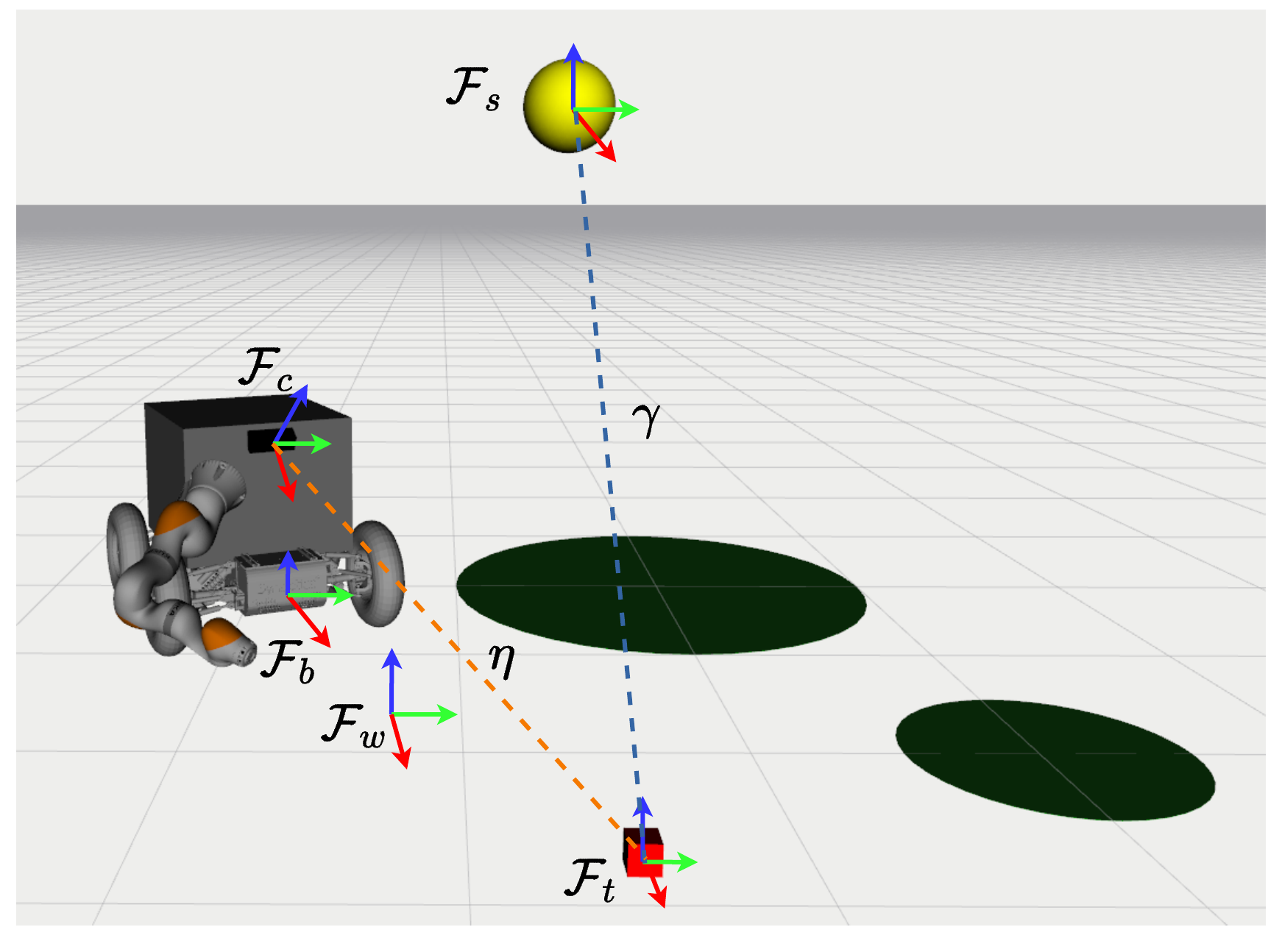}
\caption{\label{fig:ref_frames}Planning scenario with the indication of reference frames used for planning; $\eta$ is the line passing through the origins of $\mathcal{F}_c$ and $\mathcal{F}_t$, $\gamma$ that passing through the origins of $\mathcal{F}_s$ and $\mathcal{F}_t$.}
\end{figure}

In Fig.\ \ref{fig:ref_frames}, all the reference frames involved in the proposed use case are reported. $\mathcal{F}_w$ is the \emph{world} reference frame, lying on the terrain; $\mathcal{F}_b$ is the robot's \emph{base link} reference frame; $\mathcal{F}_c$ is the \emph{camera} reference frame, rigidly attached to the mobile base; $\mathcal{F}_s$ is the \emph{sun} reference frame whose relative position to $\mathcal{F}_w$ is given in terms of azimuth and elevation; $\mathcal{F}_t$ is the frame fixed with the \emph{target}.

The overshadowing avoidance constraint is verified by checking that the volume of the robot $V({\bf q}(t_f))$ at the final waypoint does not intersect the straight line $\gamma$ connecting $\mathcal{F}_s$ to $\mathcal{F}_t$: 
\begin{equation} \label{eq:shadow_constr}
    V({\bf q}(t_f)) \cap \gamma = \emptyset. 
\end{equation}

\subsection{Complete constraints formulation}

Let us define the set $\mathcal{A}(t)$, containing all joints' configurations respecting the geometric (positional) constraints, \ie: 
\begin{equation}
    \mathcal{A}(t) = \Bigr\{ {\bf q}(t): \eqref{eq:joint_pos_limits},\eqref{eq:env_collision_constraints}, \eqref{eq:self_collision_constraints}, \eqref{eq:forbidden_areas_constraint} \text{ hold} \Bigl\}.
\end{equation}
At the final waypoint, $\mathcal{A}(t_f)$ also accounts for \eqref{eq:shadow_constr}, \ie:
\begin{equation}
    \mathcal{A}(t_f) = \Bigr\{ {\bf q}(t_f): \eqref{eq:joint_pos_limits},\eqref{eq:env_collision_constraints}, \eqref{eq:self_collision_constraints}, \eqref{eq:forbidden_areas_constraint}, \eqref{eq:shadow_constr} \text{ hold} \Bigl\}.
\end{equation}

Then, the set $\mathcal{B}(t)$, employed in \eqref{eq:dp_optimization_problem} and \eqref{eq:derivative_constr}, contains the joint configurations in $\mathcal{A}$ that also allow satisfying constraints on first- and second-order derivatives, i.e.\ the reachable nodes at time $t$:
\begin{equation}
    \mathcal{B}(t) = \Bigr\{ {\bf q}(t) \in \mathcal{A}(t): \eqref{eq:joint_vel_limits},\eqref{eq:joint_acc_limits}, \eqref{eq:base_joint_limits}, \eqref{eq:pure_rolling_constr} \text{ hold}\Bigl\}.
\end{equation}
Here, to verify the differential constraints on positions, the derivatives in \eqref{eq:joint_vel_limits}, \eqref{eq:joint_acc_limits}, \eqref{eq:base_joint_limits}, \eqref{eq:pure_rolling_constr} are approximated with the Euler method, or equivalent discrete approximation.

\section{Objective Function formulation} \label{sect:objective_function}

As introduced in Section \ref{sect:introduction}, the objective function to optimize in our use case is associated to the target visibility (TV) from a base-mounted camera.
Without loss of generality, and only for the sake of simplicity, we assume that the camera is fixed during the entire trajectory, which is a typical choice for navigation/localization cameras, thus a fixed transformation $\mathcal{T}_c^b$ exists between the base and the camera frames, so that the camera position ${\bf x}_c(t)$ in the world frame $\mathcal{F}_w$ can be computed, at each time instant, from the transformation matrix
\begin{equation}
     \mathcal{T}_c^w(t) = \mathcal{T}_b^w\big({\bf q}_b(t)\big) \mathcal{T}_c^b,
\end{equation}
where $\mathcal{T}_b^w\big({\bf q}_b(t)\big)$ is the matrix transforming $\mathcal{F}_b$ into $\mathcal{F}_w$.
With reference to Fig.\ \ref{fig:ref_frames}, the cost function is formalized as the dot product between the $x$-axis of the camera frame (identified by the unit vector $\hat{\boldsymbol{x}}$) and the unit vector $\hat{\boldsymbol{\eta}}$ lying onto the line $\eta$ connecting the camera to the target:
\begin{equation}
    \zeta_{TV}\big({\bf q} \big) = 1 - \hat{\boldsymbol{x}} \cdot \hat{\boldsymbol{\eta}}.
\end{equation}
Note that, since we are solving a minimization problem, the 1-complement of the dot product is considered. In order to also penalize solutions requiring high energy consumption, which is crucial for space robots, the squared norm of joint velocities $\zeta_{SNV}$ is considered: 
\begin{equation}
    \zeta_{SNV} = \dot{{\bf q}}^T\dot{{\bf q}}.
\end{equation}
This index limits unnecessary joint movements and is often used as a simple surrogate of energy consumption when an accurate energy model is not available \cite{Kazerounian_1988}. Therefore, the objective function $\phi$ in \eqref{eq:dp_optimization_problem} and \eqref{eq:optimal_smoothing_problem} is multi-objective and is formally defined as: 
\begin{equation} \label{eq:mo}
    \phi = \sigma \zeta_{TV} + (1 - \sigma) \zeta_{SNV},
\end{equation}
where $0 \leq \sigma \leq 1$ weighs the two performance indices. 

\section{3DROCS} \label{sect:3DROCS}

In order to demonstrate that OptiWB is suitable to be employed in real mission planning scenarios, in this section, we discuss its integration with 3DROCS, a Rover Operations Control Software by Trasys International, designed to prepare, monitor and control rover planetary surface operations, with the help of a 3D graphical environment. It includes models of robotic systems and the environment in which they operate, \eg terrain models, textures, lighting and on-line shadowing, relief maps to allow the scientific and engineering ground operators to indicate and verify sites of interest and dangerous areas that are beyond the limits of the rover’s capabilities. It also provides mission-specific functions, such as Activity Plans (AP) specification, predictions related to power, communication and mechanical subsystems.

As far as planning is concerned, 3DROCS allows setting targets that are locations of scientific interest on the terrain, defining paths to be followed by the rover, and indicating forbidden areas. However, the optimality problem of the rover's motion is not addressed directly for the required paths, but the planning system relies on a simple decoupled approach where the mobile base path is planned first, followed by motion planning of the arm, when the target becomes proximal. This approach cannot guarantee optimal movements and might require complex maneuvers in the proximity of the target to satisfy some application constraints, like optimal target visibility and shadowing. In this context, we propose to integrate OptiWB with 3DROCS to fill this gap.

\begin{figure}[]
\centering
\includegraphics[width=0.41\textwidth]{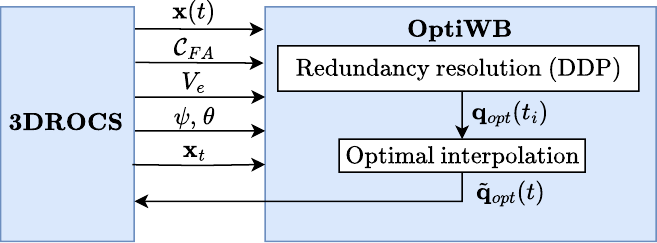}
\caption{\label{fig:3DROCS_ROS_architecture}Data interface between 3DROCS and OptiWB.}
\end{figure}

Since the concerned processes exist in the realm of offline ground planning, the integration only requires the definition of suitable data formats and modules to read and write interface data files, e.g.\ \texttt{json}. In particular, the environment description $V_e$, \ie obstacle specifications, the forbidden areas $\mathcal{C}_{FA}$, sun position in terms of azimuth $\psi$ and elevation $\theta$, the target position ${\bf x}_t$, and the task space trajectory ${\bf x}(t)$ are specified by 3DROCS. Then, OptiWB returns the optimal joint space trajectory $\tilde{\bf q}_{opt}(t)$. These interfaces are summarized in Fig.\ \ref{fig:3DROCS_ROS_architecture}.

\section{Experimental results} \label{sect:experimental_results}

\begin{figure}[]
\centering
\includegraphics[width=0.3\textwidth]{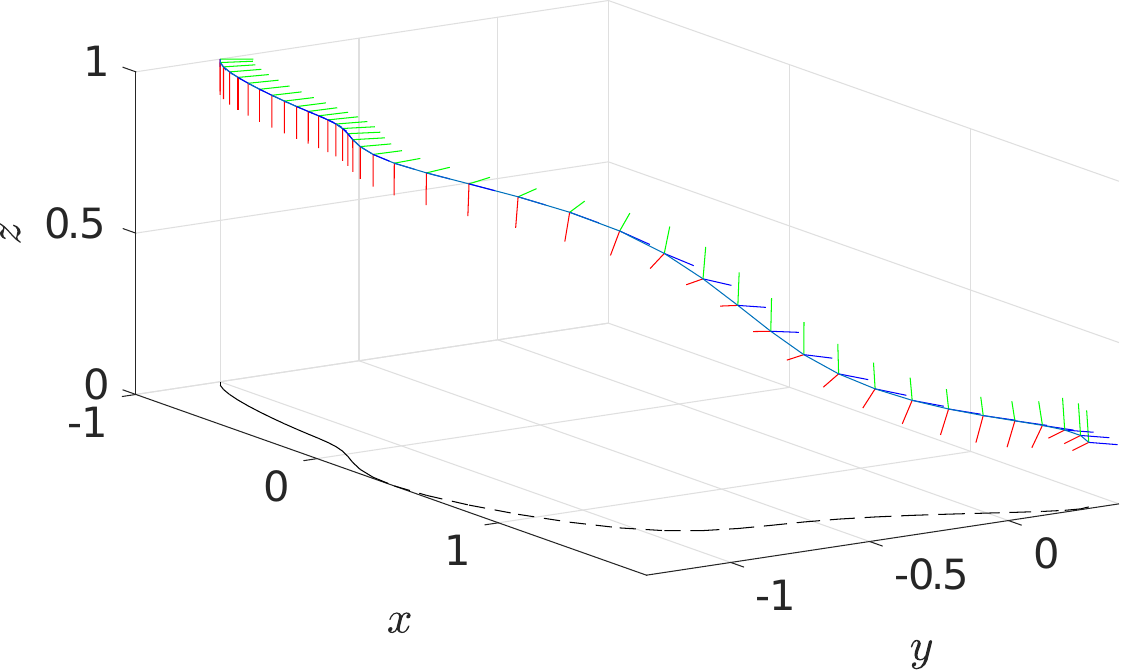}
\caption{\label{fig:workspace_traj}Assigned task space trajectory (solid, blue) and its projection on the $x$-$y$ plane (dashed, black). Frames for each waypoint represent the end-effector orientation.}
\end{figure}

Experiments are performed on a simplified version of the \textit{Analog-1 Interact rover} \cite{analog_1_interact_rover} composed of a four-wheel moving platform and a LBR iiwa 14 R820 7-DOF arm \cite{kuka} mounted on the front panel, and slightly shifted to the right with respect to the center of the panel. This configuration provides 10 DOFs, as altitude, roll and pitch of the base are assumed to be constrained by the contacts with the flat ground. For the sake of simplicity, a task with $m=7$ is assigned, yielding $r=3$. The task constrains the end-effector position, as well as one of the arm's joints, \ie the third joint. The redundancy parameters are selected to be the vector ${\bf q}_b$, as in \cite{Salvioli_2022}.
The assigned task space trajectory has 40 waypoints, a duration of $t_f=39$ s and extends for $3$ m along the $x$-axis, $1.5$ m along the $y$-axis and $0.9$ m along the $z$-axis, as shown in Fig.\ \ref{fig:workspace_traj}. The planning scenario is the one of Fig.\ \ref{fig:ref_frames}.

\begin{figure}[]
\centering
\begin{subfigure}{0.23\textwidth}
\centering
\includegraphics[width=\textwidth]{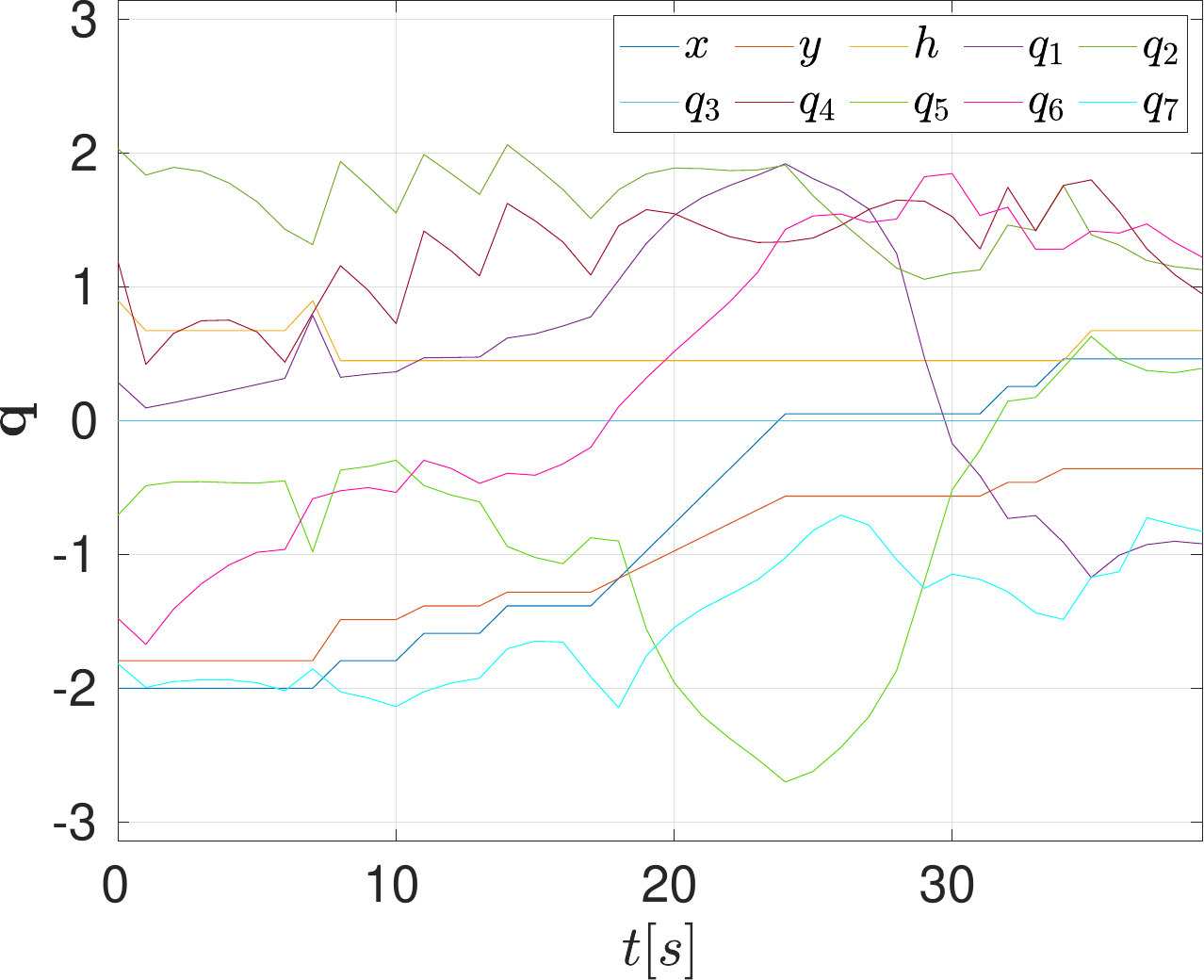}
\caption{\label{fig:joint_traj_ddp}}
\end{subfigure}
\begin{subfigure}{0.23\textwidth}
\centering
\includegraphics[width=\textwidth]{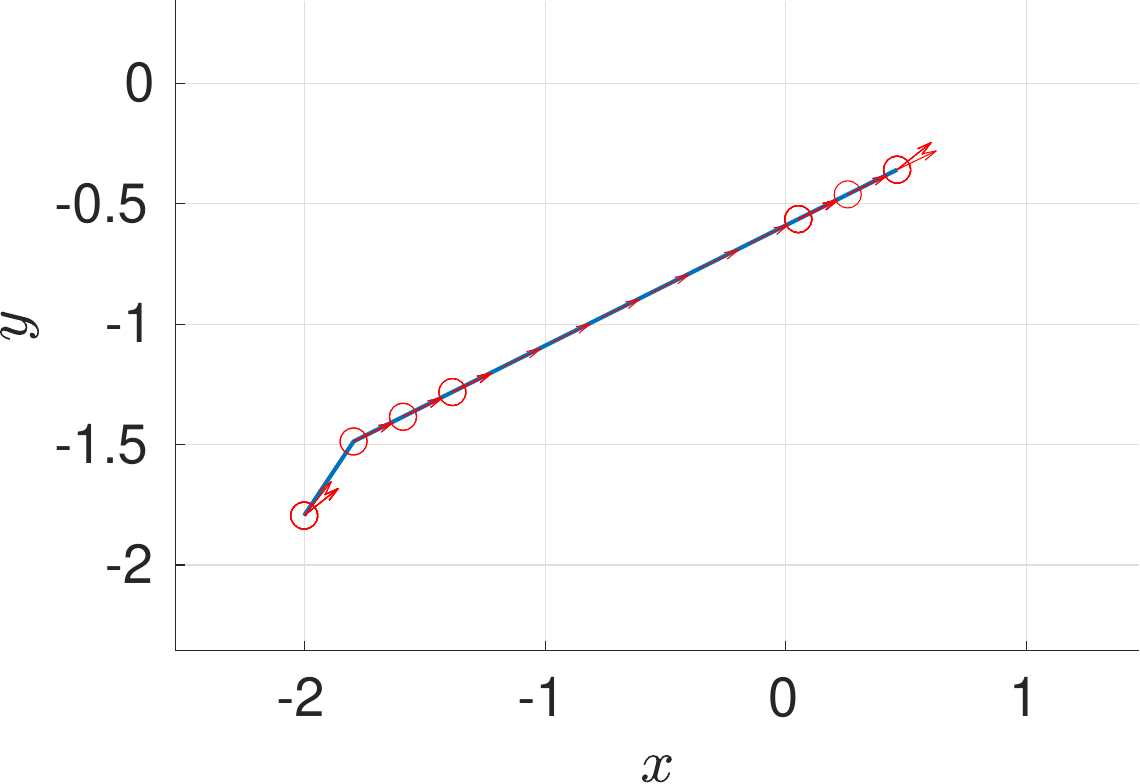}
\caption{\label{fig:base_traj_ddp}}
\end{subfigure}
\caption{\label{fig:DDP_solution} (a) Robot joint trajectory ${\bf q}(t)$ planned with DDP and (b) corresponding base path in the $x$-$y$ plane. Red arrows represent the base heading while circles represent the points where the base stops and performs a turn-in-place maneuver. }
\end{figure}

The DDP planner is implemented in ROS, while the optimal interpolation procedure is implemented in MATLAB, solving \eqref{eq:optimal_smoothing_problem} with the interior point method. The task space trajectory is inverted with the DDP algorithm using a grid resolution of $0.1$ m for $x$ and $y$, and $0.2$ rad for $h$. In \eqref{eq:mo}, $\sigma = 0.95$. The planned joint space trajectory and the corresponding base path in the $x$-$y$ plane are provided in Fig.\ \ref{fig:DDP_solution}. 

\begin{figure}[]
\centering
\begin{subfigure}{0.23\textwidth}
\centering
\includegraphics[width=\textwidth]{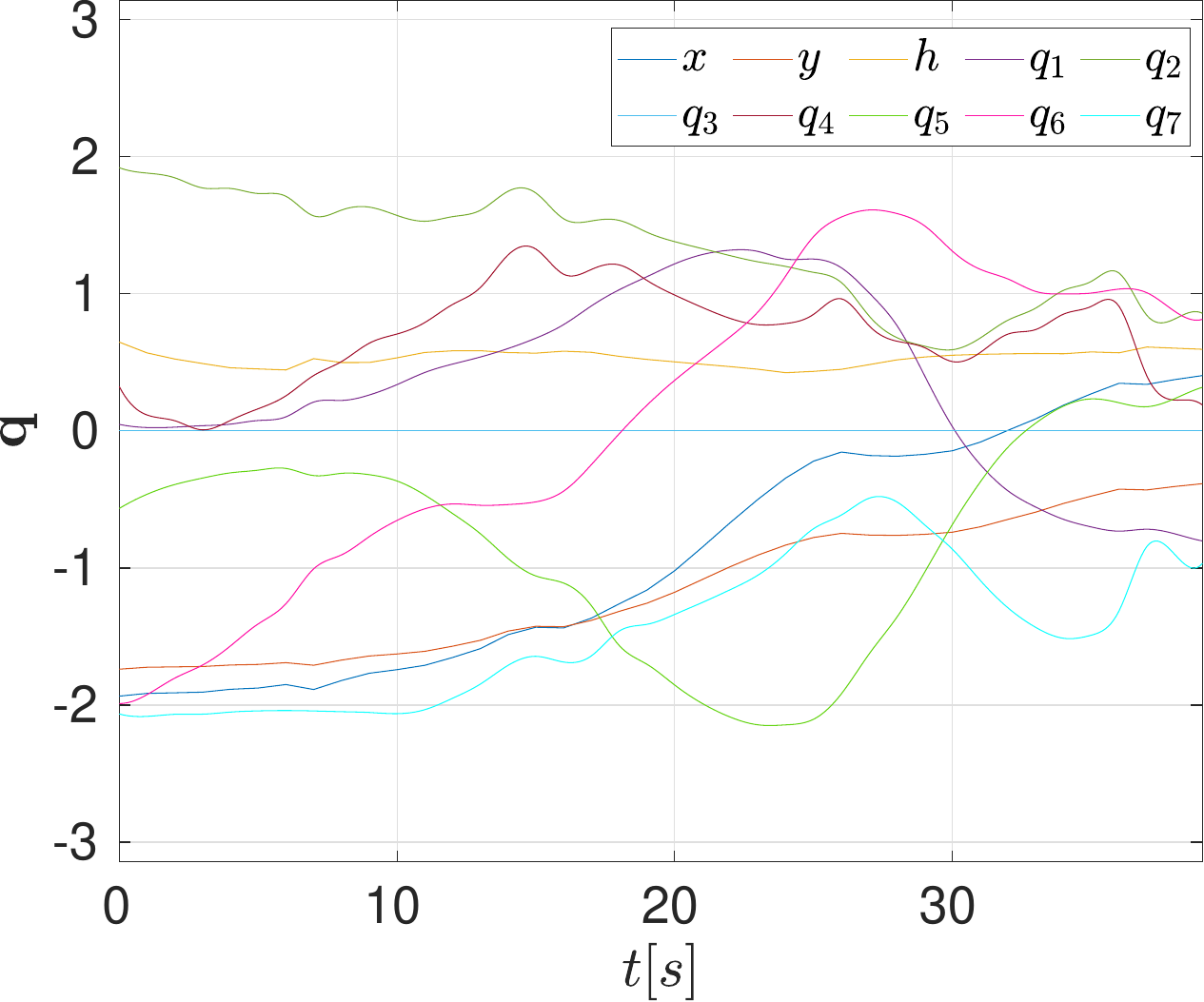}
\caption{\label{fig:smooth_joint_traj}}
\end{subfigure}
\begin{subfigure}{0.23\textwidth}
\centering
\includegraphics[width=\textwidth]{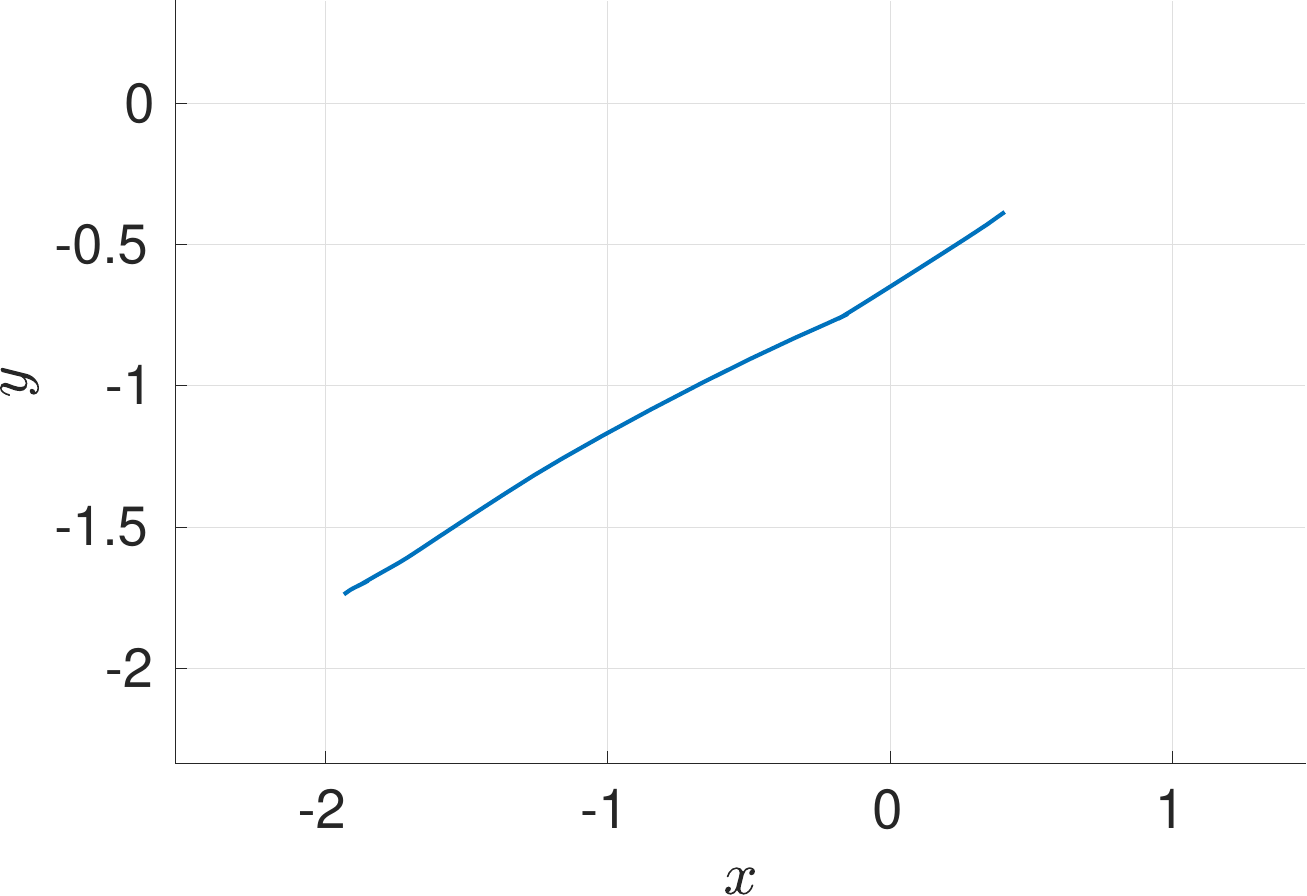}
\caption{\label{fig:smooth_base_traj}}
\end{subfigure}
\caption{\label{fig:smoothed_solution} (a) Robot joint trajectory $\tilde{\bf q}_{opt}(t)$ found after optimal interpolation and (b) corresponding base path in the $x$-$y$ plane. }
\end{figure}

The cost of the final solution is $I_{opt}(N) = 1.5829$, with integral cost for each performance index equal to $I_{TV} = 0.9695$ and $I_{SNV} = 13.2399$.
We can notice that, due to DDP discretization, the joint positions are not smooth. Also, the heading $h$ exhibits rapid changes and the base frequently stops at discrete configurations (see circles in Fig.\ \ref{fig:base_traj_ddp}). This trajectory is not suitable to generate controller references onboard, as feasibility would be likely to be compromised. Therefore, the solution ${\bf q}_{opt}(t_i)$ of DDP is used as the initial guess, i.e.\ initial spline control points, for optimal interpolation in \eqref{eq:optimal_smoothing_problem}-\eqref{eq:derivative_constr}, generating smooth (and optimal) arm and base trajectories $\tilde{\bf q}_{opt}(t)$. They are reported in Fig.\ \ref{fig:smooth_joint_traj} and Fig.\ \ref{fig:smooth_base_traj}, respectively.

The final cost of the post-processed solution is $I_{opt}(N) = 1.002$, with $I_{TV} = 0.8109$ and $I_{SNV} = 4.6332$. Comparing this cost with the intermediate one from DDP, we note that, although B-spline interpolation sets higher-order constraints (not included in the problem formulation with DDP), the cost for both performance indices decreases. In addition, as desired, the solution is smooth, which is a crucial aspect for trajectory feasibility at control level.

\begin{figure}[]
\centering
\includegraphics[width=0.26\textwidth]{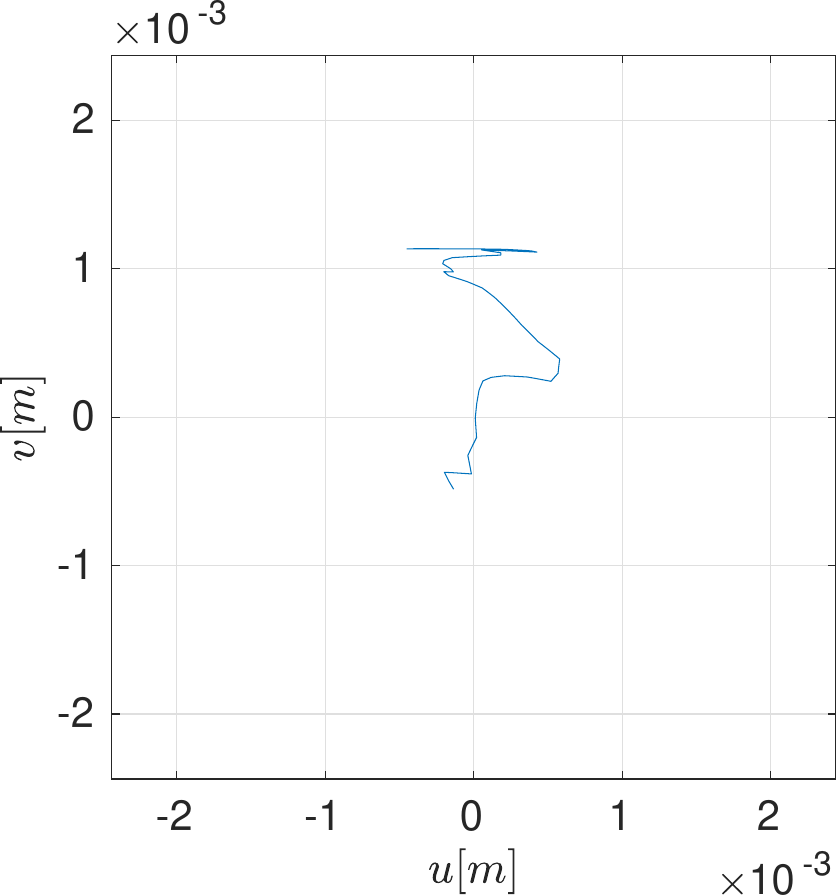}
\caption{\label{fig:obj_fcn_along_path}View of the target on the $u$-$v$ image plane when executing the interpolated trajectory. The axes are in meters and correspond to the CCD size.}
\end{figure}

The accompanying video \cite{video} shows the DDP and interpolated trajectories in RViz. At the last waypoint, the robot never stands between the sun and the target, and the base motion is planned in a way that the target is kept in the center of the camera's visual cone as much as possible during trajectory execution (Fig.\ \ref{fig:obj_fcn_along_path}). On the other hand, for the same use case, if no constraints are taken into account during planning, forbidden areas violation and target shadowing are possible. This is shown in the second section of the video: at the beginning of the trajectory, the forbidden area is traversed while, at the end, the base overshadows the target.

Once the optimal plan is available, it is returned to 3DROCS, providing a more detailed description of the environment, for rehearsal and validation. The last section of the video shows the plan execution in the 3DROCS simulator: as required, the target always lies in the camera vision cone (in blue), the robot's shadow does not cover the target and the red forbidden areas are avoided.

\section{Conclusions} \label{sect:conclusions}

Robot motion planning for space exploration is of fundamental importance during mission planning and operations. Most state-of-the-art tools used at ground control centers do not provide planning modules effectively exploiting kinematic redundancy to optimize performance indices of interest while satisfying constraints. We proposed an Optimal Whole Body Planner (OptiWB) performing a two-step optimization based on Discrete Dynamic Programming (DDP) and optimal interpolation. Focusing on a planetary mobile manipulator use case, constraints typical of space exploration are formalized, such as collision, forbidden areas, and target overshadowing avoidance. In addition, a performance index based on the target visibility from navigation cameras is considered. The effectiveness and applicability of OptiWB are demonstrated by integrating it with a commercial mission planner, 3DROCS, employed in several ESA missions. The results show that a feasible and optimal plan is generated satisfying the application constraints. At the same time, the ground operator is relieved from designing cumbersome rover and arm maneuvers: the whole-body approach simplifies the planning process, as non-trivial movements can be planned from the sole specification of a desired task-space trajectory.

We envisage to extend our planning system towards more challenging scenarios of planetary construction, where e.g., the system is characterized by many degrees of redundancy. On the technological side, we aim at integrating OptiWB within the new born Space ROS ecosystem.

\addtolength{\textheight}{-12cm}   




\bibliographystyle{IEEEtran}
\bibliography{IEEEabrv, bibliography.bib}

\end{document}